%% file: main.tex
\documentclass[10pt,conference]{IEEEtran}
\IEEEoverridecommandlockouts

\usepackage{cite}
\usepackage[normalem]{ulem}
\usepackage{graphicx}
\usepackage{subcaption}
\usepackage{multirow}
\usepackage{tabularx}
\usepackage{caption}
\usepackage{todonotes}
\usepackage{amsmath} 
\usepackage{url}

\usepackage{listings}
\usepackage{xcolor} 

\newcommand{\stitle}[1]{\vspace{0.3em}\noindent\emph{\textbf{#1}}}  

\definecolor{codegreen}{rgb}{0,0.6,0}
\definecolor{codegray}{rgb}{0.5,0.5,0.5}
\definecolor{codepurple}{rgb}{0.58,0,0.82}
\definecolor{backcolour}{rgb}{0.95,0.95,0.92}

\lstdefinestyle{mystyle}{
    backgroundcolor=\color{backcolour},   
    commentstyle=\color{codegreen},
    keywordstyle=\color{magenta},
    numberstyle=\tiny\color{codegray},
    stringstyle=\color{codepurple},
    basicstyle=\ttfamily\footnotesize,
    breakatwhitespace=false,         
    breaklines=true,                 
    captionpos=b,                    
    keepspaces=true,                 
    numbers=left,                    
    numbersep=5pt,                  
    showspaces=false,                
    showstringspaces=false,
    showtabs=false,                  
    tabsize=2
}

\definecolor{background}{HTML}{ffffff}
\definecolor{comment}{HTML}{008000}
\definecolor{keyword}{HTML}{0000FF}
\definecolor{string}{HTML}{A31515}
\definecolor{number}{HTML}{098658}
\definecolor{annotation}{HTML}{800080}
\definecolor{delimiter}{HTML}{000000}
\definecolor{variable}{HTML}{267F99}
\definecolor{function}{HTML}{AF00DB}
\definecolor{type}{HTML}{267F99}
\definecolor{punctuation}{HTML}{000000}

\definecolor{background}{HTML}{ffffff}
\definecolor{comment}{HTML}{008000}
\definecolor{keyword}{HTML}{0000FF}
\definecolor{string}{HTML}{A31515}
\definecolor{number}{HTML}{098658}
\definecolor{variable}{HTML}{267F99}
\definecolor{function}{HTML}{AF00DB}
\definecolor{annotation}{HTML}{800080}
\definecolor{delimiter}{HTML}{000000}

\lstdefinestyle{javastyle}{
    language=Java,
    basicstyle=\ttfamily,
    backgroundcolor=\color{background},
    commentstyle=\color{comment},
    keywordstyle=\color{keyword},
    stringstyle=\color{string},
    numberstyle=\color{number},
    identifierstyle=\color{variable},
    emphstyle=\color{function},
    moredelim=[is][\bfseries]{@@}{@@}, 
    morekeywords={println, int, string, void, class},
    numbers=left,
    stepnumber=1,
    numbersep=5pt,
    showspaces=false,
    showstringspaces=false,
    tabsize=4,
    frame=single,
    breaklines=true
}

\lstdefinestyle{gostyle}{
    language=Go,
    basicstyle=\ttfamily,
    backgroundcolor=\color{background},
    commentstyle=\color{comment},
    keywordstyle=\color{keyword},
    stringstyle=\color{string},
    numberstyle=\color{number},
    identifierstyle=\color{variable},
    emphstyle=\color{function},
    moredelim=[is][\color{delimiter}]{@@}{@@},
    morekeywords={Println, fmt, runtime},
    numbers=left,
    stepnumber=1,
    numbersep=5pt,
    showspaces=false,
    showstringspaces=false,
    tabsize=4,
    frame=single,
    breaklines=true
}

\lstdefinestyle{pythonstyle}{
    language=Python,
    basicstyle=\ttfamily,
    backgroundcolor=\color{background},
    commentstyle=\color{comment},
    keywordstyle=\color{keyword},
    stringstyle=\color{string},
    numberstyle=\color{number},
    identifierstyle=\color{variable},
    emphstyle=\color{function},
    moredelim=[is][\color{delimiter}]{@@}{@@},
    morekeywords={print, def, import, for, if, else, assert, frame},
    numbers=left,
    stepnumber=1,
    numbersep=5pt,
    showspaces=false,
    showstringspaces=false,
    tabsize=4,
    frame=single,
    breaklines=true
}

\newcommand{\xx}{\mathbf{x}}
\newcommand{\kk}{\mathbf{k}}
\newcommand{\val}{\mathbf{v}}
\newcommand{\myNum}[1]{(\emph{#1})}
\newcommand{\javacode}[1]{\lstinline[style=javastyle]|#1|}
\newcommand{\gocode}[1]{\lstinline[style=gostyle]|#1|}
\newcommand{\pythoncode}[1]{\lstinline[style=pythonstyle]|#1|}

\usepackage{tcolorbox}
\newcounter{findingCounter}
\begin{document}

\title{Looking into Black Box Code Language Models}




\author{
    \IEEEauthorblockN{Muhammad Umair Haider\IEEEauthorrefmark{1}, 
    Umar Farooq\IEEEauthorrefmark{2}, 
    A.B. Siddique\IEEEauthorrefmark{1}, 
    Mark Marron\IEEEauthorrefmark{1}}
    \IEEEauthorblockA{\IEEEauthorrefmark{1}Department of Computer Science, University of Kentucky\\
    Email: mha361@g.uky.edu, siddique@cs.uky.edu, marron@cs.uky.edu}
    \IEEEauthorblockA{\IEEEauthorrefmark{2}School of Electrical Engineering and Computer Science, Louisiana State University\\
    Email: ufarooq@lsu.edu}
}



\maketitle

\begin{abstract}
Language Models (LMs) have shown their application for tasks pertinent to code and several code~LMs have been proposed recently.  
The majority of the studies in this direction only focus on the improvements in performance of the LMs on different benchmarks, whereas LMs are considered black boxes. 
Besides this, a handful of works attempt to understand the role of attention layers in the code~LMs.
Nonetheless, feed-forward layers remain under-explored which consist of two-thirds of a typical transformer model's parameters.

In this work, we attempt to gain insights into the inner workings of code language models by examining the feed-forward layers. 
To conduct our investigations, we use two state-of-the-art code~LMs, Codegen-Mono and Ploycoder, and three widely used programming languages, Java, Go, and Python.
We focus on examining the organization of stored concepts, the editability of these concepts, and the roles of different layers and input context size variations for output generation.
Our empirical findings demonstrate that lower layers capture syntactic patterns while higher layers encode abstract concepts and semantics. 
We show concepts of interest can be edited within feed-forward layers without compromising code~LM performance.
Additionally, we observe initial layers serve as ``thinking'' layers, while later layers are crucial for predicting subsequent code tokens.
Furthermore, we discover earlier layers can accurately predict smaller contexts, but larger contexts need critical later layers' contributions.
We anticipate these findings will facilitate better understanding, debugging, and testing of code~LMs.


\end{abstract}



\input{intro}
\input{background}

\input{approach}

\input{keys}

\input{values}
\input{related}

\section{Conclusions}
\label{sec:conclusions}
This work targets a key problem in code~MLs -- understanding the inner workings and interpretability of code language models. 
Our study focused on feed-forward layers of LMs, which consist of two-thirds of a typical transformer model's parameters. 
In our investigations, we employ two state-of-the-art code language models, Codegen-Mono and Polycoder, and leverage three widely-used programming languages, Java, Go, and Python, as the basis for our analyses.
Our empirical findings show lower layers capture syntax while higher layers encode abstract concepts and semantics. 
We demonstrate concepts can be edited in feed-forward layers without compromising the code language model's performance. 
Initial layers serve as ``thinking'' layers, while later layers crucially predict subsequent tokens. 
Earlier layers can accurately predict smaller contexts, whereas the role of later layers becomes critical in facilitating better predictions. 
We anticipate that these findings will lay the groundwork for developing a more comprehensive understanding, enabling more effective debugging and testing methodologies for code language models.



\bibliographystyle{plain}

\bibliography{references.bib}

\end{document}

%% file: intro.tex
\section{Introduction}
\label{sec:intro}

Code language models (code LMs), leveraging the transformers architecture~\cite{vaswani2017attention}, have emerged as powerful productivity tools in software development. 
Inspired by the success of natural language processing (NLP) transformers (e.g., BERT~\cite{devlin2019bert}, GPT~\cite{radford2018improving}), these models have been trained on vast repositories of code from open-source projects.
Through this training, code LMs have acquired the ability to capture complex patterns, syntax, and semantics of programming languages.
Consequently, code LMs have demonstrated significant success across various coding tasks, including code generation, completion, editing, and documentation. Notably, many of them including GitHub Co-pilot~\cite{github-2021-copilot} and Amazon CodeWhisperer~\cite{amazon-2023-codewhisperer} are getting incorporated into integrated development environments (IDEs) as assistants to improve developers' productivity.

Existing work on code LMs, such as CodeBERT~\cite{feng2020codebert}, GraphCodeBERT~\cite{guo2020graphcodebert}, CodeGPT~\cite{codegpt}, and CodeT5~\cite{wang2021codet5} primarily focus on the performance improvement of the code LMs on different benchmarks and treat code LMs as a \emph{black box}.
Specifically, 96\% of studies focus on improving the predictive accuracy of code LMs~\cite{jiarpakdee2021practitioners}. 
These studies overlook a crucial aspect: understanding the underlying mechanisms by which these models make predictions or generate code. 
As a consequence, the inner workings of code LMs remain largely obscure, potentially resulting in the generation of vulnerable code~\cite{9833571}, challenges in debugging~\cite{huang2023large, 10.1145/3597503.3623306}, and difficulties in updating the codebase~\cite{10.1145/3586030}. 
Moreover, the lack of interpretability undermines developers' confidence in these models and their ability to effectively leverage them in practical software development scenarios. 
Enhancing the interpretability of code LMs is critical for enhancing transparency, trust, compliance, and accountability in software development.

Recognizing the importance of interpretability in code LMs, Authors in \cite{mohammadkhani2023explaining} focused on understanding the role of attention layers in code LMs.
Their study examined the distribution of attention weights across input sequences, shedding light on a crucial aspect of model behavior. 
Nonetheless, it is noteworthy that attention layers constitute only one-third of a typical code LM.
The remaining two-thirds, primarily constituted by feed-forward (FF) layers, have largely remained unexplored in existing research. 
Moreover, in NLP literature, FF layers are considered the databases (i.e., memory) of the model, represented in the form of keys and values~\cite{geva2020transformer}. 
In this work, we aim to bridge this gap by concentrating on the FF layers of code LMs, aiming to explain their role and impact in code LMs.

Specifically, for a given code prefix as input, we compute the activation coefficient for a selected key in a certain layer.
Then, we obtain the top code prefixes whose representation produced the highest inner product with the given key. 
Upon analyzing these prefixes, we discover interesting syntactic and semantic patterns associated with each key. 
Likewise, when we mask keys related to a specific concept of interest (e.g., \pythoncode{numpy}), we observe a notable decrease in the performance of the code LMs concerning that particular concept. However, other programming constructs do not exhibit significant performance deterioration following the masking of the same keys. 
Additionally, we transform each value vector into a probability distribution by multiplying it with the output embedding matrix. 
Then, we assess how the predictions at each layer align with the final output of the model. Furthermore, we manipulate the context size to investigate the impact of varying context lengths on this alignment.

In our investigation, we employ two well-known autoregressive code LMs: Codegen-Mono-2.7B~\cite{nijkamp2022codegen} and Polycoder-2.7B~\cite{xu2022systematic}. 
Codegen specializes in the Python programming language, while Polycoder encompasses multiple languages, where our focus is on three diverse programming languages: Java, Go, and Python.
To conduct our exploration, we collected 5,000 code files from active GitHub repositories with more than 50 stars for each programming language. 

Specifically, our study focuses on the following research questions (RQs).

\vspace{0.3em}
\noindent\textbf{RQ1:} \emph{What information is stored in the feed-forward layers of code LMs?}

Considering the unexplored role of FF layers in code LMs, our inquiry aims to uncover what information is stored in FF layers.
We examine the top 50 input sequences against each key in the FF layers, which exhibit the highest activation in that key relative to all other sequences in the dataset.
We then qualitatively and quantitatively explored these keys to see how the model is storing information to uncover insights into the nature of information representation in the FF layers, particularly in relation to code generation tasks.
Our investigation revealed that the FF layers of code LMs are responsible for capturing a wide range of information, spanning from fundamental syntactic patterns such as keywords and n-grams to more abstract concepts and semantics. 
Notably, the initial layers predominantly capture low-level syntactic elements (e.g., keywords, n-grams), while the higher layers capture more abstract and higher-level semantics, such as iterators and other complex programming constructs.

\vspace{0.3em}
\noindent\textbf{RQ2:} \emph{Can we precisely edit a concept of interest in code LMs, and how does such editing affect the general performance of code LMs?}

If we truly understand how information is stored in the FF layers, then we must be able to edit it.
We aim to find out the feasibility of accurately editing the concept of interest in code LMs and to evaluate the subsequent impact on the model's overall performance. 
This inquiry is motivated by the need to quantify the adaptability of code LMs to new information, particularly concerning deprecated methods or application programming interfaces (APIs).
To address this question, adopt a systematic approach. 
Initially, we identify and filter keys associated with APIs of interest, such as \pythoncode{numpy} in Python, across various programming languages using regular expressions, focusing on those keys where our concept of interest ranks among the top 50 triggers. Subsequently, we apply masking techniques to these keys and observe the effect on the model's performance concerning the concept of interest. 
Conversely, we evaluate the impact of masking the same keys on the model's performance on everything except the concept of interest, aiming to quantify any potential side effects on general performance.
Our findings indicate a significant decrease in accuracy concerning the concept of interest, implying that the model's knowledge is highly localized. 
Additionally, we did not observe a noteworthy decline in the model's performance regarding all other aspects except for the concept of interest.
This empirical evidence demonstrates the viability of editing operations without detrimentally affecting its general performance.

\vspace{0.3em}
\noindent\textbf{RQ3:} \emph{How does local information in each layer agree to the final output of code LMs?}

This question seeks to explain how the final output of the model is constructed across layers and to what extent agreement exists between different layers and the ultimate output of the model. 
This inquiry is motivated by the desire to gain insights into the flow of information in the model and to comprehend the mechanisms underlying the formulation of the model's final output.
To investigate this question, we multiply the output of each layer with the output embedding matrix, apply argmax operation to the output of each layer, and compare the top token prediction of the last layer with those of all preceding layers.  
This operation enables us to quantify the degree of agreement between different layers and assess how information is processed and consolidated throughout the model.
We observe that initial layers exhibit limited agreement with the final layers, suggesting that they primarily function as ``thinking'' layers rather than directly contributing to the output. 
In contrast, as we progress through the model, we observe an increase in agreement, indicating that later layers, which possess more processed information, play a pivotal role in generating the final output. 
This empirical evidence sheds light on the hierarchical nature of information processing in code LMs.

\vspace{0.3em}
\noindent\textbf{RQ4:} \emph{How does the context size impact the agreement between layers in code LMs?}

This question investigates the effect of context size on the agreement between layers in code LMs. 
We aim to understand how variations in context size, ranging from shorter to longer sequences, influence the degree of agreement among different layers in the model.
This inquiry is motivated by the interest in assessing how the complexity of the task for the model changes with varying context sizes.
Our analysis reveals that earlier layers can accurately predict smaller initial contexts, whereas as the context size increases, the task of predicting the correct output becomes more challenging.
Only later layers demonstrate the capability to predict accurately in such scenarios, indicating a significant impact of context size on the model's performance and problem difficulty.

\stitle{Summary of findings}.
We explore how FF layers encode syntactic and semantic information of programming languages and their role in generating output tokens in code~LMs. 
Our empirical findings demonstrate that lower layers capture syntactic patterns, while higher layers encode abstract concepts and semantics. 
We also show that concepts of interest can be edited within FF layers without compromising the performance of code~LMs. 
Additionally, we observe that initial layers serve as ``thinking'' layers, while later layers are crucial for predicting the next tokens of code. 
Furthermore, we discover that earlier layers can make accurate predictions for smaller contexts, while larger contexts pose a greater challenge, where the role of later layers is critical. 

\stitle{Contributions}.
In summary, this work makes the following contributions:

\begin{itemize}
    \item We explore and describe the role of feed-forward layers in code language models, which consist of two-thirds of a typical transformer model's parameters. 
    \item We demonstrate the viability of editing a concept of interest in code language models and empirically show the impact of editing concepts on model performance. 
    \item We explore how information aggregates through the models and the impact of context size variations and different layers on the models' final outputs.
\end{itemize}

Next, we discuss the background of this work (in Sec.~\ref{sec:background}), Sec.~\ref{sec:approach} provides details about our approach including selected code~LMs and dataset.
We present our investigations and findings on information storage and editing in Sec.~\ref{sec:keys} and 
Sec.~\ref{sec:values} provides insights and discussions on layer agreement to output and the impact of context size on code~LMs.
We discuss related work in Sec.~\ref{sec:related} and provide conclusions of this work in Sec.~\ref{sec:conclusions}.

%% file: background.tex
\section{Background}
\label{sec:background}
In this section, we discuss the necessary background on transformer-based language models, code LMs, and neural memories.

\subsection{Transformer-based Language Models}

The Transformer architecture~\cite{vaswani2017attention} employs interconnected attention blocks and feed-forward layers. 
The attention block~\cite{bahdanau2014neural} facilitates the model's ability to weigh the significance of individual tokens in a sequence, thus capturing long-range dependencies across the input sequence. 
Concurrently, the feed-forward layers enable the model to retain crucial information derived from the training data~\cite{geva2020transformer}.
The transformer-based LMs are trained using extensive text data in a self-supervised manner. 
Their substantial parameter space, often reaching billions or even trillions, gives them an impressive ability to absorb broad semantic and syntactic knowledge and strong memorization skills.
These models have achieved state-of-the-art performance for various NLP tasks, and the utilization of transformer-based LMs has emerged as a highly promising research direction in NLP~\cite{vaswani2017attention,radford2018improving,devlin2019bert,radford2019language,raffel2020exploring}.

\begin{table*}[t!]
  \centering
  \begin{tabular}{c|c|c}
    \hline
    Encoder-Decoder (e.g., T5) & Encoder (e.g., BERT) & Decoder (e.g., GPT) \\ \hline
    \includegraphics[width=0.3\textwidth]
    {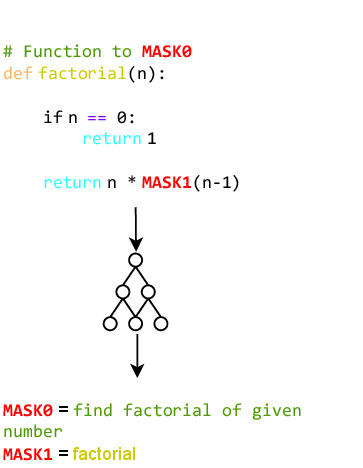} & 
    \includegraphics[width=0.3\textwidth]{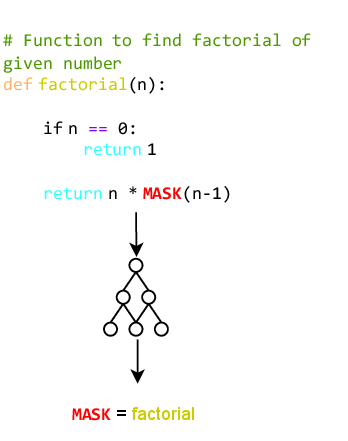} & 
    \includegraphics[width=0.3\textwidth]{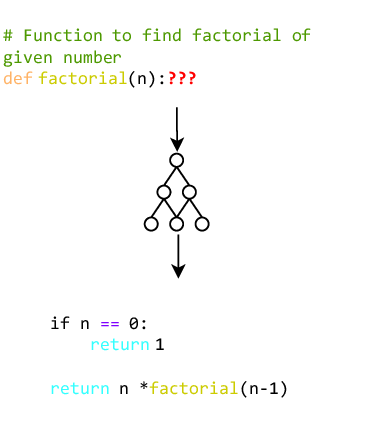} \\ 
    \hline
  \end{tabular}
  \caption{Code LMs follow transformer architecture, which comes in three variations: encoder-decoder, encoder-only, and decoder-only.}
  \label{tab:figure_table}
\end{table*}

Transformer-based LMs have three variations in their architecture.
Table~\ref{tab:figure_table} illustrates these architectures.
Encoder-decoder models, such as T5~\cite{raffel2020exploring}, adhere to the original transformer architecture, with both encoder and decoder stacks.
They formulate tasks by framing them as text-to-text problems, enabling unified training and inference.
Encoder models, such as Bidirectional Encoder Representations from Transformers~(BERT)~\cite{devlin2018bert}, utilize the encoder stack and adopt a masked language modeling objective during training. 
They leverage bidirectional context understanding to comprehend text effectively.
Decoder models, such as Generative Pre-trained Transformer~(GPT)~\cite{radford2019language}, leverage the decoder stack. 
They are trained to predict the next tokens based on preceding ones, they excel in language generation tasks. 
Due to the simplicity of the decoder architecture and the prevalence of text generation tasks, decoder models have become a de facto standard for various language modeling tasks.



\subsection{Code Language Models}

Following the success of transformer architecture in NLP, code LMs have adopted this architecture.
In code LMs, there are primarily three categories, mirroring the classifications of transformer models. These are Masked LMs, Encoder-Decoder, and Decoder-only autoregressive models.

Masked LMs in the context of coding generate code for masked tokens by classifying them based on the adjacent tokens on either side~\cite{devlin2019bert}. 
The advantage of Masked LMs over autoregressive models lies in their ability to consider the context from both sides of a masked token, providing a richer base of information for predicting the masked token. 
Examples of Masked LMs tailored for coding include CodeBert~\cite{feng2020codebert} and CuBERT~\cite{kanade2020learning}.

The predominant category in code LMs is the auto-regressive models, which focus on predicting the subsequent token based on the preceding context. 
The GPT models~\cite{gpt-1} belong to of decoder-only category and the T5 models~\cite{t5-paper} are encoder-decoder models. 
In Encoder-Decoder models an encoder encodes the input, which is then passed to a decoder akin to GPT for multiple mask prediction. 
The Code-specific Encoder-Decoder models include CodeT5~\cite{wang2021codet5} and PLBART~\cite{ahmad2021unified}.
Lastly, Decoder-Only models (i.e., GPTs) estimate the likelihood of the next token based on previous ones. 
In the broader field of NLP, GPT-like models have achieved prominence, a trend that extends to code LMs as well. 
Decoder-Only models for code feature \cite{lu2021codexglue, xu2022systematic, nijkamp2022codegen, chen2021evaluating, black2022gpt}, among others.

The widespread adoption of auto-regressive models, including GPT variants, is primarily due to their sequential left-to-right token prediction capability. 
This trait enables their application in a variety of contexts, such as code completion, generating comments for code, or converting plain text into code~\cite{feng2020codebert, guo2020graphcodebert}.

\subsection{Neural Memories}

\begin{figure}[t!]
  \centering
  \includegraphics[width=0.48\textwidth]{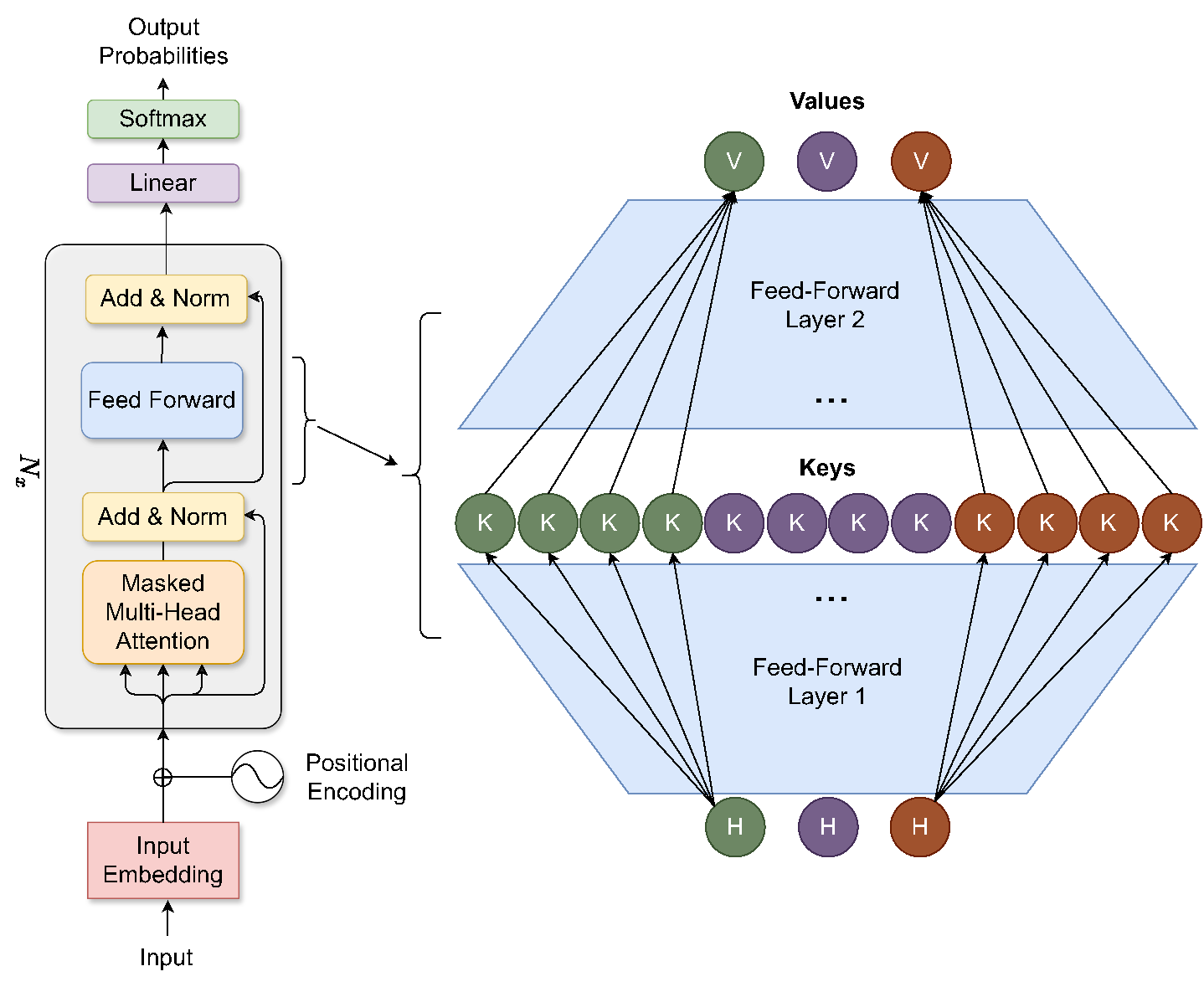}
  \caption{Feed Forward layers act as key-value memories of the model. Feed Forward layers constitute two-thirds of a typical code LM.}
  \label{fig:my_label}
\end{figure}

Authors in \cite{sukhbaatar2019augmenting} has shown that feed-forward layers act as key-value memories, emulating memory networks~\cite{sukhbaatar2015end}.
For a given input context $\xx$, we can compute the distribution over keys: $p(k_i \mid x) \propto \exp(\xx \cdot \kk_i)$ and memory of $\xx$ can be expressed as $M(\xx) = \sum_{i=1}^{d_m} p(k_i \mid x) \val_i$.
That is, we can represent FF layers as $\text{FF}(\xx) = f(\xx \cdot K^{\top}) \cdot V$, where $\xx \in \real^d$ is the text input, $K, V \in \real^{d_m \times d}$ represent parameter matrices and $f$ denotes a non-linearity~\cite{geva2020transformer}.

Code~LMs follow the transformer architecture~\cite{vaswani2017attention}, which incorporates interconnected self-attention and FF layers. 
Each FF layer operates as a position-wise function, independently processing input vectors.
The FF layers function using two matrices: one representing keys and the other values. 
The first matrix serves as a set of key vectors, while the second matrix serves as a set of corresponding values for these keys.
Specifically, transformers employ ReLU non-linearity and the function of FF layers can be expressed as: 
$\text{FF}(\xx) = \text{ReLU}(\xx \cdot K^{\top}) \cdot V,$ where $\xx$ represents the input vector, $K$ represents the output of the first matrix acting as keys, and $V$ represents the output of the second matrix acting as values.
Figure~\ref{fig:my_label} illustrates the zoomed-in view of FF layers, emphasizing the keys and values.

%% file: approach.tex
\section{Approach}
\label{sec:approach}
In this section, we discuss our approach to conducting our study, including selected code models, dataset, and research questions. 

\subsection{Selected Models}

For our choice of models, we chose two state-of-the-art mid-sized models for our investigation. 
One is a mono-language model and the other is a multi-language model.

\stitle{Codegen-Mono-2.7B.} Codegen\cite{nijkamp2022codegen} a 2.7 billion parameter GPT model, with 32 layers, it is trained sequentially on three datasets, called \textsc{ThePile}~\cite{gao2020pile}, \textsc{BigQuery}~\cite{bigquery},
and \textsc{BigPython}. 
\textsc{ThePile} is an 825.18 GB English text corpus for language modeling. 
The dataset is constructed from 22 diverse high-quality subsets, one of which is programming language data collected from GitHub repositories with more than 100 stars that constitute 7.6\% of the dataset. 
The multi-lingual dataset \textsc{BigQuery} is a subset of Google’s publicly available BigQuery dataset, which consists of code in multiple programming languages. 
For the multi-lingual training, the following 6 programming languages are chosen: C, C++, Go, Java, JavaScript,
and Python. 
The monolingual dataset \textsc{BigPython} contains a large amount of data in the Python programming language. 

\stitle{Polycoder-2.7B.} Polycoder~\cite{xu2022systematic} is also a 2.7 billion parameter GPT model, with 32 layers, it was trained on cloned repositories for 12 popular programming
languages with at least 50 stars (stopping at about 25K per language to avoid a too-heavy skew towards popular programming languages) from GitHub in October 2021. 
For each project, each file belonging to the majority language of that project was extracted, yielding the initial training set. 
This initial, unfiltered dataset spanned 631GB and 38.9M files.

\begin{table}[t!]
\begin{tabular}{|c|c|c|c|c|}
\hline
\textbf{Language} & \textbf{\begin{tabular}[c]{@{}c@{}}Number \\ of\\ Files\end{tabular}} & \textbf{\begin{tabular}[c]{@{}c@{}}Number of\\ Repositories\end{tabular}} & \textbf{\begin{tabular}[c]{@{}c@{}}Minimum\\ number of\\ stars\\ (GitHub)\end{tabular}} & \textbf{\begin{tabular}[c]{@{}c@{}}Date Last\\ Active\\ (GitHub)\end{tabular}} \\ \hline
Python            & \multirow{3}{*}{5,000}                                                 & \multirow{3}{*}{50}                                                       & \multirow{3}{*}{50}                                                                     & \multirow{3}{*}{01-01-2020}                                                    \\ \cline{1-1}
Go           &                                                                       &                                                                           &                                                                                         &                                                                                \\ \cline{1-1}
Java              &                                                                       &                                                                           &                                                                                         &                                                                                \\ \hline
\end{tabular}
\caption{Criteria for dataset collection.}
\label{tab:criteria}
\end{table}

\subsection{Dataset}
We leverage GitHub~\cite{github} to access publicly available source code, which hosts a wide array of programming languages and diverse projects. 
To establish a comprehensive dataset, we systematically cloned the most prominent repositories associated with three popular programming languages; Java, Go, and Python~\cite{top-pls}.
Our selected programming languages are representative of popular programming paradigms, imperative, dynamic, and object-oriented.
Moreover, the open ecosystem in these programming languages allows us to be selective while collecting dataset to maintain high quality.  
To maintain the quality, we selected repositories with a minimum of 50 stars (similar to Polycoder~\cite{xu2022systematic}).
We curate 5,000 files for each of the selected programming languages, Table~\ref{tab:criteria} presents the criteria we imposed while collecting the dataset from Github repositories.

Table~\ref{tab:Dataset_Details} provides a summary of the characteristics of our dataset.
In terms of source lines of code, Python, Go, and Java each one has over 1.4M, 2.1M and 572K, respectively. 
Python files exhibit over 38K classes and 93K functions, while Go files has 23K \gocode{struct} counts (Go does not have class) and 101K functions defined, and in Java files, there are 15K classes containing 27K methods.

\begin{table}[t!]
\begin{tabular}{|c|c|c|c|c|c|}
\hline
\textbf{Language} & \textbf{\begin{tabular}[c]{@{}c@{}}Number \\ of Source\\ Lines\end{tabular}} & \textbf{\begin{tabular}[c]{@{}c@{}}Average\\ File Size\\ (Lines)\end{tabular}} & \textbf{\begin{tabular}[c]{@{}c@{}}Average\\ Number \\ of tokens\\ (per line)\end{tabular}} & \textbf{\begin{tabular}[c]{@{}c@{}}Number\\ of\\ Classes\end{tabular}} & \textbf{\begin{tabular}[c]{@{}c@{}}Number\\ of\\ Functions\end{tabular}} \\ \hline
Python            & 1,493,445                                                                      & 298.68                                                                         & 15.54                                                                                       & 38,687                                                                  & 93,076                                                                    \\ \hline
Go           & 2,194,788                                                                      & 438.95                                                                         & 12.83                                                                                       & 23,386                                                                  & 101,163                                                                   \\ \hline
Java              & 572,825                                                                       & 114.56                                                                         & 13.15                                                                                       & 15,570                                                                  & 27,671                                                                    \\ \hline
\end{tabular}
\caption{Statistics of the dataset we used for our study, number of lines of source code ranges from 572K to 2.1M, with 15K to 28K classes (\gocode{struct} is used for Go).}
\label{tab:Dataset_Details}
\end{table}

\subsection{Research Questions}
We consider the following research questions for our study on selected code LMs using our dataset described above. 

\begin{itemize}
    \item [RQ1:] \emph{What information is stored in the feed-forward layers of code LMs?}
    Given the unexplored nature of the role of FF layers in code~LMs, Our investigation aims to clarify the precise information stored within these layers.
    Considering the nature of programming languages, we want to explore how syntactic information and semantics are stored in different code~LMs. 

    \item [RQ2:] \emph{Can we precisely edit a concept of interest from code LMs, and how does such editing affect the general performance of code LMs?}
    Often in programming languages and frameworks, certain methods or APIs are deprecated and code~LMs would need to adapt to the changes.
    We explore the possibility of updating the learned concept by editing the concept of interest from code LMs. 
    Along with the possibility of editing, we also want to measure the performance impact of the editing performed. 

    \item [RQ3:] \emph{How does local information in each layer agree to the final output of code LMs?}
    The capability to generate output stands as a fundamental strength of code~LMs. 
    Our objective is to investigate how this output is formulated and delineate the distinct roles of various layers in this process.

    \item [RQ4:] \emph{How does the context size impact the agreement between layers in code LMs?}
    We want to understand the role of context size and its impact on producing output.
    This research question is driven by the desire to evaluate how the complexity of the model's task evolves with changes in context size.
\end{itemize}

The next two sections (Sec.~\ref{sec:keys} and \ref{sec:values}) explore FF layers in selected code~LMs to answer these questions and discuss the findings of our study.

%% file: keys.tex
\section{Information storage and editing }
\label{sec:keys}
The following section describes our methods and experiments to find out how information is stored in FF layers~(RQ1), how can we edit stored concepts, and the impact of editing on code~LMs~(RQ2).  

\subsection{Information Storage}

\subsubsection{Capturing Top Trigger Examples.}
Let us denote our dataset as $D$, which consists of $n$ code prefixes represented as $\{\xx_1, \xx_2, \cdots, \xx_n\}$.
A code prefix $\xx_i$ is passed through the model and an activation coefficient $a_i = \text{max}(x^l_i \cdot k^l_i)$ is computed for every key $\kk^l_i$ in layer $l$, where $x^l_i$ denotes the representation of $\xx_i$ at layer $l$, and $k^l_i$ is the key vector corresponding to the $i$-th hidden dimension at layer $l$.
This process is repeated for all the prefixes in $D$.
Then a ranking of $\xx \in D$ is established for each key $\kk^l_i$ based on the activation coefficient $a$.
For each key $\kk^l_i$ in layer $l$, we then identify $t$ trigger examples $\{\xx_1, \xx_2, \cdots, \xx_t\} \subset D$, which produce activation coefficients that rank in the top 50 of a particular key $\kk^l_i$.



Authors in \cite{geva2020transformer} suggest that these keys act as detectors for specific patterns from the input data. 
By examining top-t triggers for a key, we can deduce what patterns that key is responsive to. 
This method of probing allows us to uncover the encoded patterns in a given code~LM's keys. 
That is, we can discover how the model encodes and interprets information and the model's operational logic.
%


\begin{table*}[t!]
\centering

\small
\begin{tabular}{l|l|l}
\hline
Key & Pattern & Triggers \\
\hline
\multirow{3}{*}{$k^{1}_{1}$} & \multirow{3}{*}{Keyword \pythoncode{frame}} & \pythoncode{frame_ab.shape[2]} \\
                      &                           & \pythoncode{start_frame_num = start_frame} \\
                      &                           & \pythoncode{os.path.join(pred_frame_path,} \\
\hline
\multirow{3}{*}{$k^{7}_{1}$} & \multirow{3}{*}{Keyword \pythoncode{assert}} & \pythoncode{self.assertEqual(self.buffer.read(), original)} \\
                      &                           & \pythoncode{self.assertIsNone(ret.exception)} \\
                      &                           & \pythoncode{self.assertIsNotNone(getattr(ctx.obj, name))} \\
\hline
\multirow{3}{*}{$k^{11}_{4}$} & \multirow{3}{*}{Slicing in python} & \pythoncode{weights = weights[:, 1:1 + P - 2]} \\
                      &                           & \pythoncode{lrs[:, -5:-4, :, :]} \\
                      &                           & \pythoncode{priors[:, 2:])} \\

\hline
\multirow{3}{*}{$k^{14}_{3187}$} & \multirow{3}{*}{math} & \pythoncode{np.sin(pi * x / 2) + np.finfo(np.float32).eps)} \\
                      &                           &  \#$|B - A*W|^2$ \\
                      &                           & \pythoncode{m = np.max(np.abs(covmean.imag))} \\
\hline
\multirow{3}{*}{$k^{26}_{18}$} & \multirow{3}{*}{Image related concepts} & \pythoncode{rgb = color.lab2rgb(lab.astype(np.float64))} \\
                      &                           &  \pythoncode{elif isinstance(pic, np.ndarray):} \\
                      &                           & \pythoncode{low resolution photo of the \{\}.} \\
\hline
\multirow{3}{*}{$k^{32}_{2}$} & \multirow{3}{*}{Loss concepts} & \pythoncode{g_vggloss *= self.lambda_vgg} \\
                      &                           &  \pythoncode{= math.ceil(math.log(sr_factor,1 / self.scale_factor))} \\
                      &                           & \pythoncode{(l_d_real + l_d_fake).backward()} \\
\hline
\end{tabular}
\caption{Sample trigger examples for Python from Codegen model.}
\label{tab:python_codegen}
\end{table*}

\subsubsection{Pattern Analysis using Regular Expression Filtering}

Once we have successfully gathered top-k triggers for all the keys, we face the challenge of dealing with an extensive search space, which makes obtaining meaningful quantitative results a daunting task. 
For example, both the models under investigation in this work (i.e., Codegen-Mono, and Polycoder) are autoregressive models with 32 layers and 2560 hidden dimensions containing a total of $2,560 \times 4 \times 32 = 327,680$ keys.
Navigating through this vast expanse is no small feat.

To tackle this issue, we employ a strategic approach. We implement regular expression (regex) filtering, targeting various application programming interfaces (APIs) such as \pythoncode{numpy} and \pythoncode{torch}, as well as fundamental programming concepts like loops and conditionals. This process helps us narrow down the search space, focusing on keys related to our areas of interest within the expansive search space. 

For all of the explorations on keys, we extensively use regex filtering, along with other heuristics that are based on the frequency of occurrence of our concept of interest amongst the top triggers of each key (e.g., in key 5 out of all 50 triggers, 40 are related to \pythoncode{numpy}), accordingly, to handle the wast search space, and get meaningful insights.

\subsubsection{Qualitative Analysis of keys}
We conduct a qualitative analysis across all layers to examine the patterns of information triggered by the keys using a subset of chosen keys from each layer.
The number of provided interpretations in the paper is just a small sample of many examples observed in our analysis. The point of drawing objective conclusions from manual observations of keys is valid. In the following, we briefly describe our method to shed some light on the matter. 
We collected the top 50 trigger contexts for all 327,680 keys. We then filtered these contexts using regular expressions. In the case of NumPy, we used `r”np.”` regular expression to find the keys where the use of NumPy was amongst the top 50 triggers. (Similar regular expressions were used for other APIs.) This reduces our search space to 114,020 keys which is still huge. To further reduce this search space in each layer of the model, we organized the keys into 5 ranges based on the frequency of the occurrence of API in the top 50 triggers. Finally, we randomly selected 5 keys per layer from each range for analysis. This filtering process gives us a representative subset of 15-25 keys per layer per API, which results in a total of 480 to 800 keys per API per language and model settings. 
We show a subset of representative keys in the paper. In general, we see the same emergent behaviors as discussed in the paper.

To get a better representation of chosen keys than random, we filter the keys for a particular API, for example \pythoncode{numpy}.
We then divide the filtered keys into five different ranges based on the frequency of occurrences of the concept of interest and select five randomly selected keys from each range.
In theory, it should give us 25 keys per layer for each concept of interest, which is not always the case because frequencies of occurrence of the concept of interest are not uniformly distributed across all ranges.
Nonetheless, we manually go through approximately 15-25 keys per layer for each concept of interest.
Doing this gives us a heuristic to get a better representation of the vast search space.

We present results from some of these selected keys in Tables~\ref{tab:python_codegen}, \ref{tab:GO_polycoder}, \ref{tab:Java_polycoder}, and \ref{tab:python_polycoder}. 
We showcase a few examples from each key, specifically, we highlight three instances that exemplify the main pattern observed among the 50 triggers for that key. We have also included the original text files associated with the keys' triggers to ensure completeness.

\begin{table*}[t!]
\centering
\small
\begin{tabular}{l|l|l}
\hline
Key & Pattern & Triggers \\
\hline
\multirow{3}{*}{$k^{1}_{4}$} & \multirow{3}{*}{Keyword \gocode{runtime}} & \gocode{runtime, _ := getTestModuleInstance(t)} \\
                      &                           & \gocode{return r.runtime.regExpExec(execFn, r, s)} \\
                      &                           & \gocode{if runtime.GOOS ==}\\
\hline
\multirow{3}{*}{$k^{8}_{24}$} & \multirow{3}{*}{Keyword \gocode{func}} & \gocode{func (*AnyValue_StringValue) isAnyValue_Value()} \\
                      &                           & \gocode{func PrintToPDF() *PrintToPDFParams \{\}}\\
                      &                           & \gocode{func newBaseGoCollector() baseGoCollector}  \\
\hline
\multirow{3}{*}{$k^{15}_{7142}$} & \multirow{3}{*}{\shortstack[l]{Longer then token\\ length string}} & \gocode{func (t *Transport) time.Duration (original)} \\
                      &                           & \gocode{NativeHistogramMinResetDuration time.Duration}\\
                      &                           & \gocode{Timeout time.Duration}  \\
\hline
\multirow{3}{*}{$k^{22}_{24}$} & \multirow{3}{*}{Setting Flag Values} & \gocode{SYS_PPOLL = 336} \\
                      &                           &  \gocode{DLT_LINUX_LAPD = 0xb1} \\
                      &                           & \gocode{TCP_BBR_PACE_PER_SEC = 0x43e} \\
\hline
\multirow{3}{*}{$k^{26}_{2}$} & \multirow{3}{*}{\shortstack[l]{Checks and errors\\ on internet services}} & \gocode{conn, err := net.DialUDP("udp", nil, udpAddr)} \\
                      &                           &  \gocode{expectedFilenameURL: \&url.URL\{Scheme: "file", Path: ""\},}\\
                      &                           & \gocode{ConnectionTimeout: s.opts.connectionTimeout,}
 \\

\hline
\multirow{3}{*}{$k^{30}_{22}$} & \multirow{3}{*}{Comments} &  \gocode{/* For block sizes below 64 kB, we never need } \\
                      &                           &  \gocode{// Invariant: we have a 4-byte match at s, and } \\
                      &                           & \gocode{// before the CAS operation. So, we need to check} \\
\hline
\end{tabular}

\caption{Sample trigger examples for Go from Polycoder model.}
\label{tab:GO_polycoder}
\end{table*}

The quantitative analysis reveals that the initial layers, or lower layers, of the model, predominantly focus on identifying keywords and n-grams, such as the example frame in key 1 layer 1 (i.e., $k^{1}_{1}$) of Codegen-Mono on python in Table~\ref{tab:python_codegen}. Progressing deeper into the model, the layers exhibit an enhanced semantic understanding. 
A notable example is key 3187 in layer 14 (i.e., $k^{14}_{3187}$) of Codegen-Mono on python in Table~\ref{tab:python_codegen}, which demonstrates the model's capability not only to cluster similar mathematical functions like \pythoncode{np.math} and \pythoncode{np.sin} but also to recognize a comment that contains a mathematical equation, despite it not being code. This progression underscores a significant increase in the model's semantic understanding.
As we move further into the higher layers, the model's ability to grasp and interpret complex semantic concepts, including but not only limited to loss, image, math, and slicing, becomes increasingly apparent. Through these select examples, it is clear that the model evolves to understand higher-level semantic concepts with greater depth as we ascend through its layers.

\stitle{Analysis of Python in Codegen-Mono.}
Table~\ref{tab:python_codegen} provides triggers for Python the Codegen-Mono model, in the first two examples it is shown that the keys are capturing keywords: frame in key 1 layer 1 (i.e., $k^{1}_{1}$) and keyword assert in key 1 layer 7 (i.e.;$k^{7}_{1}$), after these initial layers it is predominantly higher level semantics, key 4 layer 11 (i.e., $k^{11}_{4}$) is about slicing of arrays in python, though it very well might be just capturing the sign \pythoncode{:}, key 3187 layer 14 (i.e., $k^{14}_{3187}$) as mentioned above is an interesting one because the model seems to group concepts of maths in this key, even equation comments are in that key. Key 18 in layer 26 (i.e., $k^{26}_{18}$) captures concepts of image, from RGB to resolution to checking if the instance is an image, the understanding of the model is so profound about concepts related to images that in a key, which we have not showcased here, it even captured an array initialization of [0,255], without any mention of the image at all. 
Key 2 layer 32 (i.e., $k^{32}_{2}$) captures concepts of loss in deep learning in Python. In the given triggers it captures from loss weightage to backpropagating loss to a manual equation of a loss function with no mention of loss keyword.

\stitle{Analysis of Go in Polycoder.}
In Table~\ref{tab:GO_polycoder}, we present triggers for the Go on the Polycoder model. In the first two rows are examples of model capturing keywords: \gocode{runtime} in key 4 layer 1 (i.e., $k^{1}_{4}$) and func in key 24 in layer 8 (i.e., $k^{8}_{24}$). In the next row, we show a key 7142 layer 14 (i.e., $k^{14}_{7142}$) that is not necessarily capturing semantics but is capturing a longer string \gocode{time.Duration} which is not just a single keyword. In the next row, the key 24 layer 22 (i.e., $k^{22}_{24}$) captures the setting of different flags with hex values. The next key 2 layer 26 (i.e., $k^{26}_{2}$) is interesting as it captures checks and errors specifically on internet services, from exceptions on file name URL to connection error and timeout, it is interesting that this key not only knows about checks and errors but also checks and errors specifically on internet services. Lastly, the key 22 layer 30 (i.e., $k^{30}_{22}$) is capturing different comments, even with different styles of commenting too (i.e., \gocode{//} or \gocode{/*}), from this key, it is evident that the model knows the difference between comments and code.

\begin{table*}[t!]
\centering
\small
\begin{tabular}{l|l|l}
\hline
Key & Pattern & Triggers \\
\hline
\multirow{3}{*}{$k^{1}_{3}$} & \multirow{3}{*}{Keyword \javacode{has}} & \javacode{hasRouterField = true;} \\
                      &                           & \javacode{hasLeadership = false; }\\
                      &                           & \javacode{hasFields = writeIfNotEmpty(out,}\\
\hline
\multirow{3}{*}{$k^{8}_{35}$} & \multirow{3}{*}{\shortstack[l]{errors and\\ exceptions}} & \javacode{ log.error("Trying to reap: " + holder.path, e);} \\
                      &                           & \javacode{System.err.println("syntax error "": " + command);}\\
                      &                           & \javacode{Assert.fail("Expected auth exception was not thrown");}  \\
\hline
\multirow{3}{*}{$k^{15}_{30}$} & \multirow{3}{*}{Time} & \javacode{long start = System.nanoTime();}\\
                      &                           & \javacode{put("nano", System.nanoTime());}\\
                      &                           & \javacode{Assert.assertTrue(listener.await(10, TimeUnit.SECONDS)}  \\
\hline
\multirow{3}{*}{$k^{21}_{3}$} & \multirow{3}{*}{Internet Protocols} & \javacode{URL url = new URL(tcpUrl.replaceFirst("tcp", "http"));} \\
                      &                           &  \javacode{return localInetAddress.getHostAddress();} \\
                      &                           & \javacode{ftpFtpConnection.ftp.changeWorkingDirectory(..);} \\
\hline
\multirow{3}{*}{$k^{26}_{4347}$} & \multirow{3}{*}{Logs and errors} & \javacode{LOG.info(String.format("-----", count));} \\
                      &                           &  \javacode{throw new IOException(String.format("Incorrect version}\\
                      &                           & \javacode{log.error(String.format("Connection timed out,}
 \\

\hline
\multirow{3}{*}{$k^{31}_{3}$} & \multirow{3}{*}{Loops} &  \javacode{for (Element mb : members) \{} \\
                      &                           &  \javacode{i < constructorBean.parameterTypes.size(); i++) \{} \\
                      &                           & \javacode{while (mc.find()) \{} \\
\hline
\end{tabular}

\caption{Sample trigger examples for Java from Polycoder model.}
\label{tab:Java_polycoder}

\end{table*}

\stitle{Analysis of Java in Polycoder.}
Table~\ref{tab:Java_polycoder} presents triggers for Java on the Polycoder model. The first row is a key 2 layer 1 (i.e., $k^{1}_{3}$) that captures the keyword \javacode{has}, and is not different from the other experiment tables, 
but the next key 35 layer 8 (i.e., $k^{8}_{35}$) is different from the previously discussed tables as it seems to have a higher level of semantic understanding of errors and exceptions in Java, from logging the error to asserting and printing errors. Given this key is not in the first few layers, it is not unexpected to capture semantics, but considering other examples where keys in this range of layers were mostly capturing keywords, it is an interesting result, showcasing that the boundary of where semantic understanding of the model starts is not super clear. 
Next key 30 layer 15 (i.e., $k^{15}_{30}$) is capturing instances of time. 
Next key 3 layer 21 (i.e., $k^{21}_{3}$) captures concepts of network connections and network protocols specifically from FTP to TCP to the local host (i.e., connections), in contrast to key 2 layer 26 (i.e., $k^{26}_{2}$) in Table~\ref{tab:GO_polycoder} which was also capturing network services, but it was specifically capturing errors and logs.
This shows the understanding of the model in different semantics. 
Next is a key 4347 layer 26 (i.e., $k^{26}_{4347}$) with logs and errors from throw to logging of info and errors. 
Next key 3 layer 32 (i.e., $k^{31}_{3}$) is unique in the sense that it captures an actual programming concept of loops.

\begin{table*}[t!]
\centering

\begin{tabular}{l|l|l}
\hline
Key & Pattern & Triggers \\
\hline
\multirow{3}{*}{$k^{1}_{4}$} & \multirow{3}{*}{Keyword \pythoncode{add}} & \pythoncode{add\_tokens=True}\\
                      &                           & \pythoncode{add(values)}\\
                      &                           & \pythoncode{add\_image\_summaries=True}\\
\hline
\multirow{3}{*}{$k^{8}_{246}$} & \multirow{3}{*}{Keyword \pythoncode{randn}} & \pythoncode{np.random.randn(10,) * 0.1} \\
                      &                           & \pythoncode{= self.rng.randn(}\\
                      &                           & \pythoncode{rnn['Bin'] = rng.randn(N)/np.sqrt(1.0)}  \\
\hline
\multirow{3}{*}{$k^{15}_{131}$} & \multirow{3}{*}{Load and Save} & \pythoncode{plt.savefig(f)}\\
                      &                           & \pythoncode{test\_labels = np.load(file\_obj)}\\
                      &                           & \pythoncode{pickle.dump(data, f)} \\
\hline
\multirow{3}{*}{$k^{22}_{17}$} & \multirow{3}{*}{Datasets} & \pythoncode{datasets.random\_mlp(5, 1000), 100)} \\
                      &                           &  \pythoncode{dset.CIFAR10(args.data\_path,transform=train\_transform)} \\
                      &                           & \pythoncode{dataset = datasets.EMPTY\_DATASET} \\
\hline
\multirow{3}{*}{$k^{26}_{2788}$} & \multirow{3}{*}{Labels} & \pythoncode{labels = np.array([], dtype=bool)} \\
                      &                           &  \pythoncode{groundtruth = np.array([], dtype=bool)}\\
                      &                           & \pythoncode{targets = np.zeros([batch\_size, num\_steps], np.int32)}
 \\

\hline
\multirow{3}{*}{$k^{31}_{5533}$} & \multirow{3}{*}{\shortstack[l]{Declarations \\with arrays}} &  \pythoncode{expected\_y\_min = np.array([3.0, 14.0], dtype=float)} \\
                      &                           &  \pythoncode{ Pixels = np.zeros((2 * d, 2 * d, 2), dtype=np.int32)} \\
                      &                           & \pythoncode{labels = tf.constant([1, 2], dtype=tf.int32)} \\
\hline
\end{tabular}

\caption{Sample trigger examples for Python from Polycoder model.}
\label{tab:python_polycoder}
\end{table*}

\stitle{Analysis of Python in Polycoder.}
The triggers for Python on the Polycoder model are presented in Table~\ref{tab:python_polycoder}. First, two rows are examples of the model capturing keywords: key 4 layer 1 (i.e., $k^{1}_{4}$) for \pythoncode{add} and key 246 layer 8 (i.e., $k^{8}_{246}$) for \pythoncode{randn} which is in line with the findings in other settings. 
Next key 131 layer 15 (i.e., $k^{15}_{131}$) captures codes for saving and loading different types of objects. Next key 17 layer 22 (i.e., $k^{22}_{17}$) captures codes for dataset initializations of different types. 
Next is a key 2788 layer 26 (i.e., $k^{26}_{2788}$) which captures labels for training. 
In the next row, we show a key 5533 layer 31 (i.e., $k^{31}_{5533}$) that captures different types of array declarations.
In all of the higher-level semantic keys, keywords are rarely repeated among different triggers, which proves that these keys are actually capturing the said higher-level concepts and are not just capturing keywords.

\begin{table*}[t!]
\centering

\begin{tabular}{l|l}
\hline
Key  & Triggers \\
\hline
\multirow{3}{*}{$k^{18}_{1187}$} & \pythoncode{labels = np.frombuffer(buf, dtype=np.uint8).astype(np.int32)} \\
& \pythoncode{euler\_angles = np.asarray(euler\_angles,dtype=np.float32)} \\
& \pythoncode{array\_frombytes(buffer, data)} \\
\hline

\multirow{3}{*}{$k^{29}_{1265}$} & \pythoncode{data = np.frombuffer(buf, dtype=np.uint8)}
 \\
& \pythoncode{uint8image: a [height, width, depth]} \\
& \pythoncode{class EditProfileViewTest(TestCase):} \\
\hline

\multirow{3}{*}{$k^{30}_{770}$} & \pythoncode{+= 1 - np.array(self.env.dones)}
 \\
& \pythoncode{x = np.round(xyt[:,[0]]).astype(np.int32)} \\
& \pythoncode{logvar.set\_shape(size\_\_xz)} \\
\hline

\end{tabular}
\caption{Polysemantic Keys with trigger examples in Codegen (Python).}
\label{tab:Polysementic_python}
\end{table*}

\stitle{Polysemantic Keys.}
In NLP interpretability literature, the concept of polysemous keys is recognized~\cite{fan2024evaluating}.
Polysemantic keys are unique in their ability to engage in the representation of multiple, often unrelated, concepts or functions. 
Unlike their counterparts that encode singular, straightforward functions, these neurons showcase a multifaceted nature, showing a more complex and interconnected representation within the model.

Interpreting what individual neurons/keys in a neural network are doing is a daunting task, exacerbated by the complexity of polysemantic neurons. Interpretability methods aim to map these neurons' functions, striving to demystify the model's internal mechanisms. 
However, the polysemantic nature of some neurons adds a significant layer of complexity, as these neurons do not adhere to the simplicity of encoding a single function or concept.

\stitle{Polysemantic keys in Codegen-Mono for Python.}
In our exploration, we also come across these polysemantic keys in coding models.
In Table~\ref{tab:Polysementic_python}, we present examples of some polysemantic keys for Python on the Codegen-Mono model. 
The first row shows a key 1187 layer 18 (i.e., $k^{18}_{1187}$) which is capturing labels, array from bytes, and euler\_angles, all of these do not belong to any one concept so it is evident that this key is not learning a singular function, instead it is a polysemantic key. The next row shows key 1265 layer 29 (i.e., $k^{29}_{1265}$) which contains examples of data from the buffer, a class declaration, and comments about an image, these triggers also do not have any common theme so this key is also polysemantic. Next key 770 layer 30 (i.e., $k^{30}_{770}$) also tells a similar story of being polysemantic.

\stitle{Findings.}
This qualitative analysis aids in revealing the nature of the patterns and semantics captured by the code~LMs. 
It enables us to observe the extent to which the model comprehends various high-level semantics, such as grouping a mathematical equation with math operations or capturing an array ranging from 0 to 255 within a key associated with image-related functions.
This analysis answers our first research question of uncovering the underlying nature of the stored data in FF layers.
We notice a consistent pattern in the information stored in FF layers. 
Specifically, the initial layers of the model tend to predominantly capture keywords, while higher layers tend to capture higher-level semantics.

\subsection{Editing Concept of Interest}
To answer our next research question about the possibility of editing a concept of interest from the model and how the editing will affect the performance of the model (RQ2), we perform the following experiments. 
In this work, we focus on a special case of editing: masking. 

\subsubsection{Masking}
The first step to mask keys related to the concept of interest is to identify these keys across all layers.
To do this, we filter through top-t triggers for each key $\kk^l_i$ in layer $l$ using regex, and identify the keys that are related to the concept of interest, among all 327,680 keys in the model. 
We mark a key as a key $\kk^l_i$ as a key related to a concept of interest only if the concept of interest (e.g., \pythoncode{numpy}) is used amongst the top-t triggers of that key.

After identifying the keys that are related to the concept of interest, we can mask them by zeroing out the weights of the key. 
That is, we set $k^l_i = 0$ if the key has been identified as a key related to the concept of interest, in the previous step.
Zeroing out weights is a known strategy to remove parts of the model, since zeroing out weights results in that key or part of the model not taking part in the model's output formation~\cite{haider2021comprehensive}.

\subsubsection{Performance on concepts of Interest}
To gauge the performance of the models on concepts of interest, we use 10,000 lines of code from our curated dataset for each language and model setting, containing concepts of interest, to perform this experiment.
We first select two highly used APIs or functions from each language, and then we filter the keys with top triggers for these functions or APIs using regex.

In the case of Python and Go we see the model's performance on generating the next token right after the \pythoncode{API.} call.
An example for \pythoncode{numpy} in Python would be the performance of the model to produce the right method after the \pythoncode{np.} (e.g., context is \pythoncode{val = np.}, ground truth is \pythoncode{array}).
In the case of Java, there is no \pythoncode{API.} type of calls so we went with the prediction of actual method names. 
An example would be \javacode{System.out.println(mystr} as context and \javacode{.equals(} as the ground truth. 
Exact regexes used for all the APIs, and functions are shown in the Table~\ref{tab:abilation} column ``API~of~Interest''.
We make sure that the selected filtered examples remain consistent between both, masked and unmasked, experiments.

In Table~\ref{tab:abilation}, accuracies are reported for ``API of Interest'' in column Baseline, where we provide performance accuracies of the unmasked models (i.e., unchanged pre-trained model) on concepts of interest, while in column Masked we provide performance accuracies of the masked models (i.e., model keys related to the concept of interest are masked by the masking technique discussed above) on concepts of interest, along with the drop in accuracy from baseline unmasked experiment. 

\stitle{General Performance.}
To gauge the general performance of models, excluding selected concepts of interest, we check the model's performance on the next token prediction on 10,000 lines of code in each setting.
10,000 lines of code are filtered from the dataset through regex to not have the concepts of interest used in any of them.

Table~\ref{tab:abilation} also reports results for ``concepts of non-Interest'', in column Baseline, where we provide general performance accuracies of the unmasked models(i.e., unchanged pre-trained model), and in column Masked we provide general performance accuracies of the masked models(i.e., model keys related to the concept of interest are masked by the masking technique discussed above), along with the drop in accuracy from baseline unmasked experiment.

\begin{table*}[t!]
\centering
\begin{tabular}{|cc|c|c|c|c|c|}
\hline
\multicolumn{2}{|c|}{\multirow{2}{*}{\begin{tabular}[c]{@{}c@{}}\textbf{Model}\end{tabular}}}  & \multirow{2}{*} {\begin{tabular}{c}
\textbf{Name of}
 \\ \textbf{API of Interest}
\end{tabular}} & \multicolumn{2}{c|}{\textbf{API of Interest}}   & \multicolumn{2}{c|}{\textbf{Concepts of non-Interest}} \\ \cline{4-7} 
\multicolumn{2}{|c|}{} &  
\multicolumn{1}{c|}{} & \textbf{Baseline} & \textbf{Masked}      & \textbf{\begin{tabular}[c]{@{}c@{}} Baseline\end{tabular}} & \textbf{\begin{tabular}[c]{@{}c@{}} Masked\end{tabular}} \\ \hline
\multicolumn{1}{|c|}{\multirow{2}{*}{\begin{tabular}[c]{@{}c@{}}CodeGen\\ Mono-2B\end{tabular}}} & \multirow{2}{*}{Python} & \pythoncode{np.}                & 61.06             & 41.07 \textcolor{orange}{$\downarrow$ \textbf{19.99}} & 61.53                                                             & 58.10 \textcolor{orange}{$\downarrow$ \textbf{3.43}}                                             \\ \cline{3-7} 
\multicolumn{1}{|c|}{}                                                                           &                         & \pythoncode{torch.}             & 59.32             & 48.36 \textcolor{orange}{$\downarrow$ \textbf{10.96}} & 61.74                                                             & 60.70 \textcolor{orange}{$\downarrow$ \textbf{1.04}}                                             \\ \hline
\multicolumn{1}{|c|}{\multirow{6}{*}{\begin{tabular}[c]{@{}c@{}}Polycoder \\ 2.7B\end{tabular}}} & \multirow{2}{*}{Python} & \pythoncode{np.}                & 55.19             & 41.26 \textcolor{orange}{$\downarrow$ \textbf{13.93}} & 80.18                                                             & 76.18 \textcolor{orange}{$\downarrow$ \textbf{4.0}}                                              \\ \cline{3-7} 
\multicolumn{1}{|c|}{}                                                                           &                         & \pythoncode{torch.}             & 54.61             & 34.77 \textcolor{orange}{$\downarrow$ \textbf{19.84}} & 79.92                                                             & 77.38 \textcolor{orange}{$\downarrow$ \textbf{4.0}}                                              \\ \cline{2-7} 
\multicolumn{1}{|c|}{}                                                                           & \multirow{2}{*}{Go}     & \gocode{log.}               & 69.23             & 62.13 \textcolor{orange}{$\downarrow$ \textbf{7.10}}   & 71.52                                                             & 70.60 \textcolor{orange}{$\downarrow$ \textbf{0.92}}                                             \\ \cline{3-7} 
\multicolumn{1}{|c|}{}                                                                           &                         & \gocode{time.}              & 67.23             & 35.42 \textcolor{orange}{$\downarrow$ \textbf{31.81}} & 71.52                                                             & 64.56 \textcolor{orange}{$\downarrow$ \textbf{6.96}}                                             \\ \cline{2-7} 
\multicolumn{1}{|c|}{}                                                                           & \multirow{2}{*}{Java}   & \javacode{.equals(}           & 75.59             & 63.09 \textcolor{orange}{$\downarrow$ \textbf{12.5}}  & 79.91                                                             & 77.87 \textcolor{orange}{$\downarrow$ \textbf{2.04}}                                             \\ \cline{3-7} 
\multicolumn{1}{|c|}{}                                                                           &                         & \javacode{.get(}              & 47.67             & 23.52 \textcolor{orange}{$\downarrow$ \textbf{24.15}} & 79.77                                                             & 68.87 \textcolor{orange}{$\downarrow$ \textbf{10.9}}                                             \\ \hline
\end{tabular}
\caption{Making results indicate that masking keys associated with the API of interest notably degrades the performance of models specifically for that API. However, the overall performance of the models across all other constructs is not significantly affected.}
\label{tab:abilation}
\end{table*}

\stitle{Findings.}
The results in Table~\ref{tab:abilation} help us answer RQ2, which is about the inquiry of the effects of precise editing in the network keys for a particular concept of interest. 
A notable drop in the model's performance can be seen for the concept of interest when the keys related to that concept are masked. 
Moreover, there was no significant decrease in the model's performance in areas other than the concept of interest. 
This provides empirical proof that it is possible to make editing changes without adversely impacting the overall performance of the model.
This finding suggests that the model's knowledge is localized, and the keys we are identifying to be related to a concept of interest are indeed related to that concept. 
This also proves that precise editing of the model's knowledge is plausible. 
Nonetheless, one might wonder why the performance drops drastically but does not completely diminish. 
There are multiple factors contributing to this phenomenon.
\myNum{i}~We only select the top 50 triggers, which implies that we deactivate a small percentage of keys in total. 
Intuitively, the performance should not have dropped to zero for the API of interest.
\myNum{ii}~We did not mask polysemantic keys, which are capable of learning multiple functions.
Masking these keys could potentially lead to unintended consequences on the model's overall performance. Further exploration in this direction is left to future research. 
Previous studies have also underscored that polysemous keys present a considerable challenge for model editing~\cite{fan_neuroneval_neurips23}.


%% file: values.tex

\section{Information Aggregation}
\label{sec:values}

This section elaborates on our approach and experiments to investigate the alignment between local information at different layer levels and the final output (RQ3) and study the effects of varying context sizes on these alignments (RQ4).

\begin{figure*}[t!]
    \centering
    \begin{subfigure}{.45\textwidth}
        \centering
        \includegraphics[width=\linewidth]{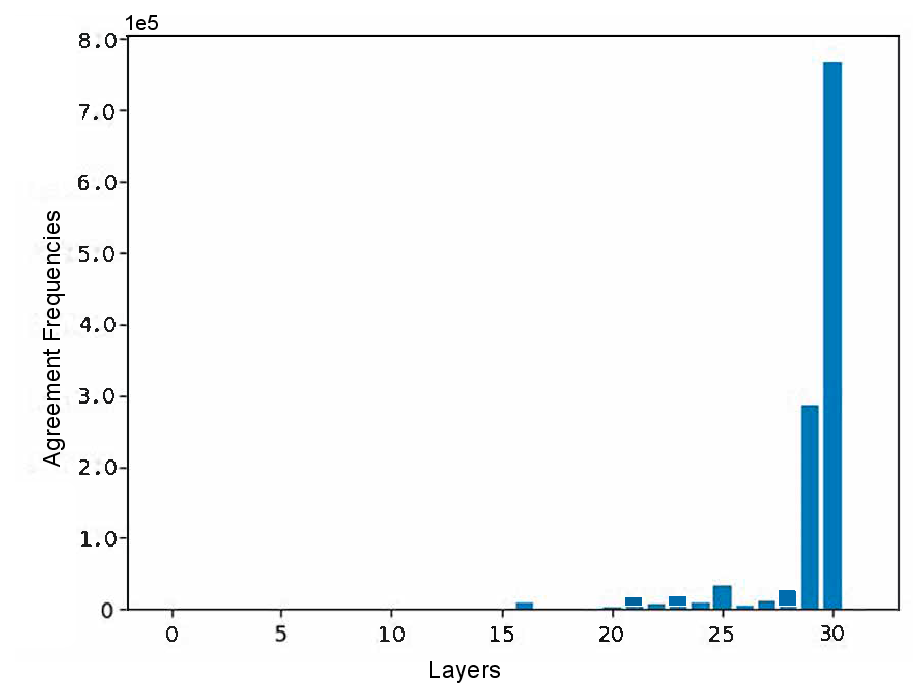}
        \caption{Python: Codegen-Mono model.}
        \label{fig:sub1}
    \end{subfigure}
    \begin{subfigure}{.45\textwidth}
        \centering
        \includegraphics[width=\linewidth]{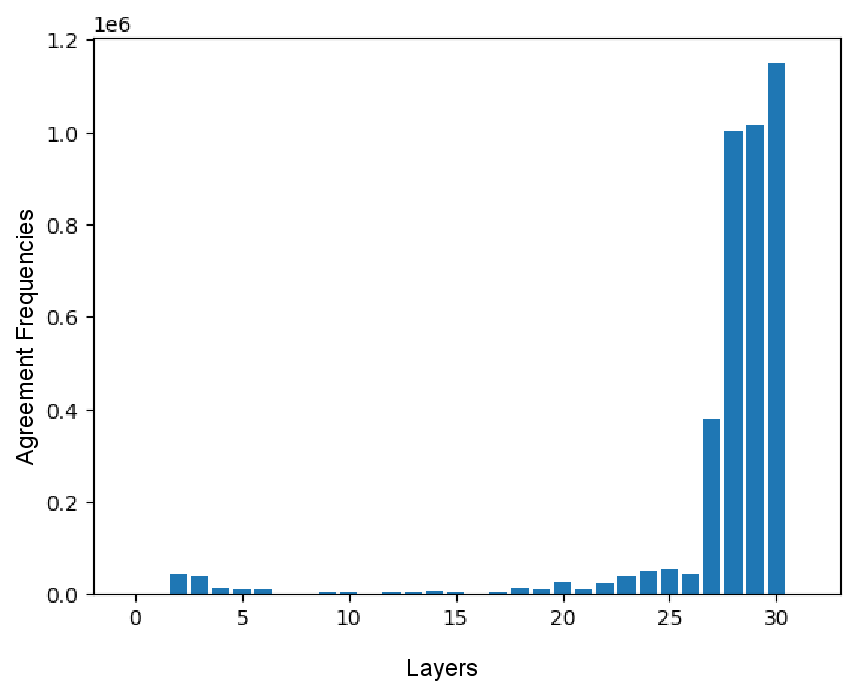}
        \caption{Python: Polycoder model.}
        \label{fig:sub2}
    \end{subfigure}
    \begin{subfigure}{.45\textwidth}
        \centering
        \includegraphics[width=\linewidth]{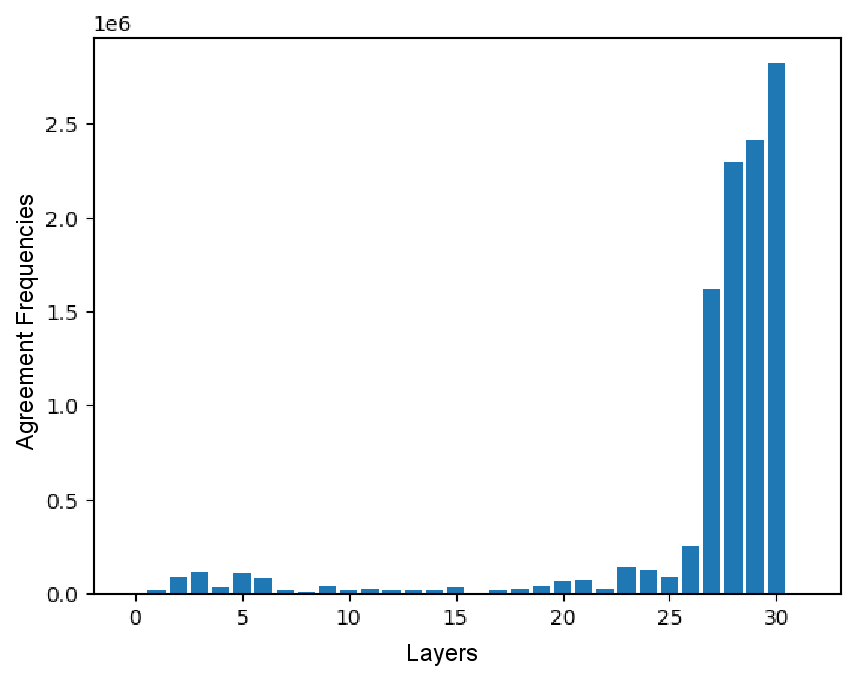}
        \caption{Go: Polycoder model.}
        \label{fig:sub3}
    \end{subfigure}
    \begin{subfigure}{.45\textwidth}
        \centering
        \includegraphics[width=\linewidth]{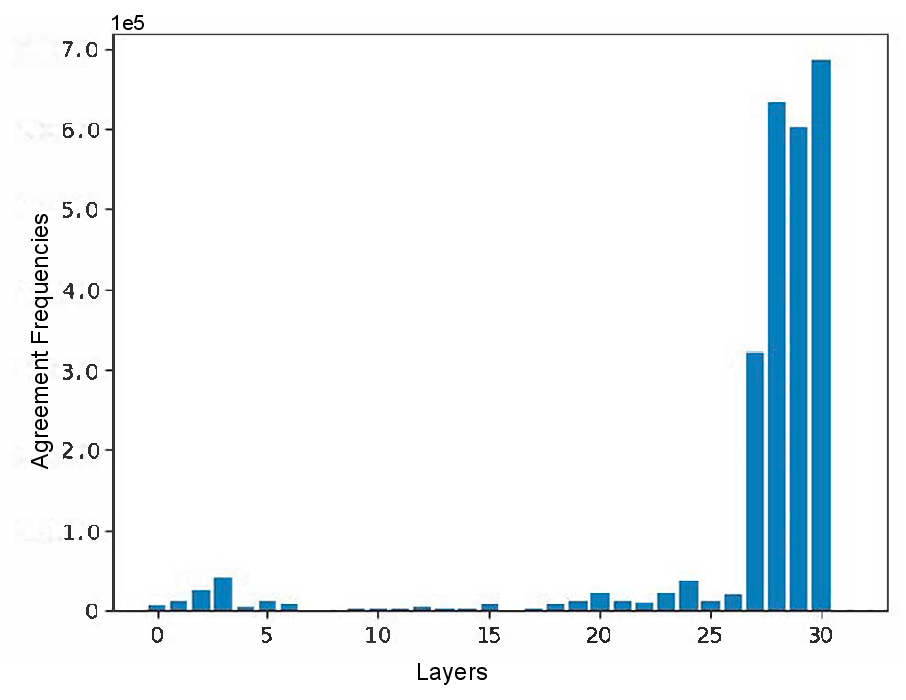}
        \caption{Java: Polycoder model.}
        \label{fig:sub4}
    \end{subfigure}

\caption{Agreement between different layers and model's final output.}
\label{fig:agreement1}

\end{figure*}

\subsection{Layer Agreements to Final Output}

To understand how different layers aggregate information to form the model's final output and whether different layers agree with the model's final output, we conduct the following experiment.
We transform each value vector (i.e., hidden dimension output of the second feed-forward layer), denoted as \(v^l_i\) in layer $l$, into a probability distribution over the vocabulary and select the token with the highest probability.
That is, we perform multiplication of \(v^l_i\) for each layer $l$ with the output embedding matrix of the model \(E\), and subsequently applying softmax function: \(p^l_i = \text{softmax}(v^l_i \cdot E)\). 
We then apply argmax function $o^l_i = \text{argmax}(p^l_i)$ to get $o^l_i$ which is the top predicted token by layer $l$, when $x_i\in D$ is passed as input to the model.


It is important to note that the resulting probability distribution \(p^l_i\) is not calibrated.
However, it is worth mentioning that the ranking established by \(p^l_i\) remains unaffected, allowing for meaningful analysis. 
To compute agreement we compare the top token prediction $o^l_i$ from each layer $l$ with the final output of the model $o^L_i$, where $L$ represents the last layer.
If $o^l_i = o^L_i$, then layer $l$ agrees with the model's final output when $x_i\in D$ is passed as input to the model.

To conduct the agreement experiment, we utilize our entire dataset. 
For each line of code, we generate multiple examples by considering all prefixes of the line, resulting in n examples, where n represents the number of tokens in the line of code.
%
Figure~\ref{fig:agreement1} presents the results of this experiment, where it is evident that the agreement of initial layers in all the settings is quite low but as we move ahead into the model the agreement starts to increase, and in the last few layers it is exponentially high.

\stitle{Python on Codegen-Mono.}
In Figure~\ref{fig:agreement1}~(a), we present agreement results for the Codegen-Mono model on Python language. 
From the graph of frequencies, it can be seen that the initial half layers of the model till layer 15 have no agreement with the final output of the model, at layer 16 there is some agreement, but it drops again till layer 20 after layer 20 it gradually increases till 28 to 29 layer, and then we see a sudden exponential increase in the agreement till the second last layer of the model.

\stitle{Python on Polycoder.}
We present agreement results for the Polycoder model on Python language in Figure~\ref{fig:agreement1}~(b). In these results, we see a little different agreement pattern where there is a little agreement in the initial layers till layer 5 then it goes down but does not become completely zero. 
After layer 20 it starts to gradually increase and after layer 25 it increases exponentially and is quite high in the last few layers of the model. 
This behavior is different from the one we noticed in the previous results in Figure~\ref{fig:agreement1}~(a), but is consistent with all the other results on the Polycoder model.
We believe that this behavior is dependent on the model and is influenced by the nature of their respective training processes, with one being monolingual and the other multilingual.

\stitle{Go on Polycoder.}
Figure~\ref{fig:agreement1}~(c) presents agreement results for the Polycoder model on Go Language. 
It shows a similar story to the agreement graph of the Polycoder model on Python language in Figure~\ref{fig:agreement1}~(b), there is some agreement in the initial layers till layer 7, then it drops but never goes to zero, then after layer 20 it gradually increases and after layer 25 it exponentially increases, and the peak of last few layers is close to each other, unlike Codegen-Mono model on Python.

\stitle{Java on Polycoder.}
In Figure~\ref{fig:agreement1}~(d) agreement results for the Polycoder model on Java language are presented. These results are similar to the other results of the Polycoder model on other languages. 
There is some agreement in the initial layers, then it drops till layer 20 and after layer 20 it gradually increases and in the last 4 layers it is exponentially high, and the peaks for the last layers are closer to each other.

\stitle{Findings.}
The results in Figure~\ref{fig:agreement1} help us to answer RQ3, how local information in each layer agrees with the final output of the code~LMs.
We observe that the early layers of the model show minimal agreement with the model's final output, implying that their primary role is akin to processing or ``thinking'' rather than having a direct impact on the output. 
Conversely, as we move deeper into the model, there is a noticeable rise in agreement, implying that the later layers, with more refined information, are more important in forming the final output.

We also observe a difference in behaviors between the two models where results for the Polycoder model have some agreement in the initial layers, which then drops and goes up again after layer 20, this is in contrast to the result for the Codegen-Mono model where initial half of the layers have no agreement with the final output of the model. 
We also observe that the peaks on high agreement in the last few layers in the Polycoder model are closer to each other, this is in contrast to the Codegen-Mono model where the peaks in the last layers are also exponential to each other. We posit that it is a model-dependent behavior and has to do with the nature of training of these two models, one being monolingual while the other being multilingual.



\subsection{Impact of Variance in Context Size to Layer Agreements}

\begin{figure*}[t!]
    \centering
    \begin{subfigure}{.45\textwidth}
        \centering
        \includegraphics[width=\linewidth]{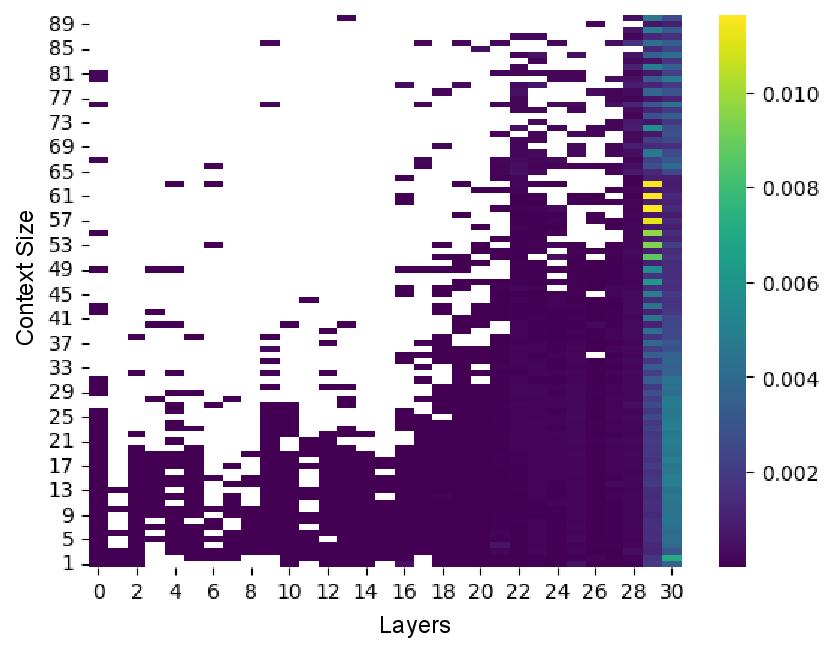}
        \caption{Python: Codegen-Mono model.}
        \label{fig:sub21}
    \end{subfigure}%
    \begin{subfigure}{.45\textwidth}
        \centering
        \includegraphics[width=\linewidth]{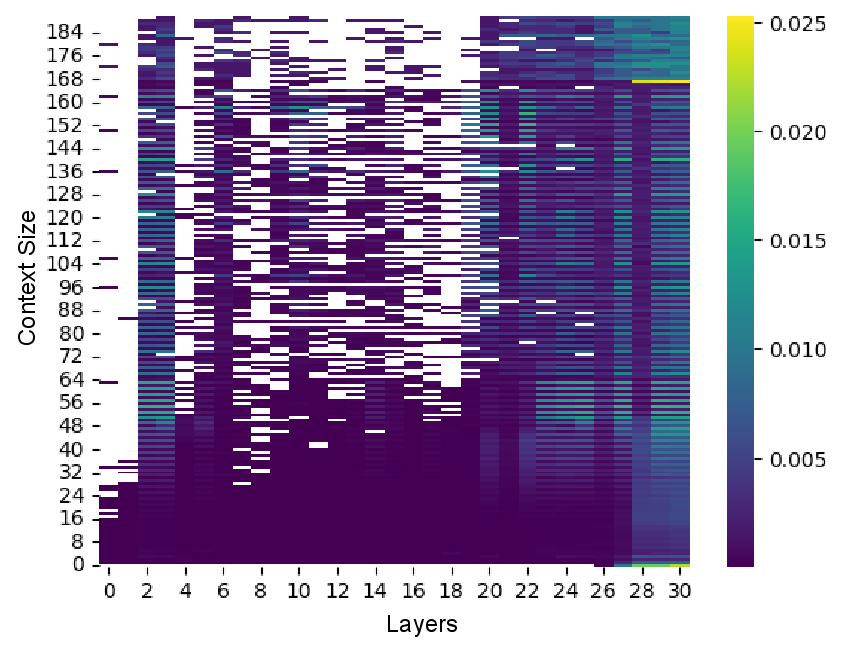}
        \caption{Python: Poly-Coder model.}
        \label{fig:sub22}
    \end{subfigure}
    
    \begin{subfigure}{.45\textwidth}
        \centering
        \includegraphics[width=\linewidth]{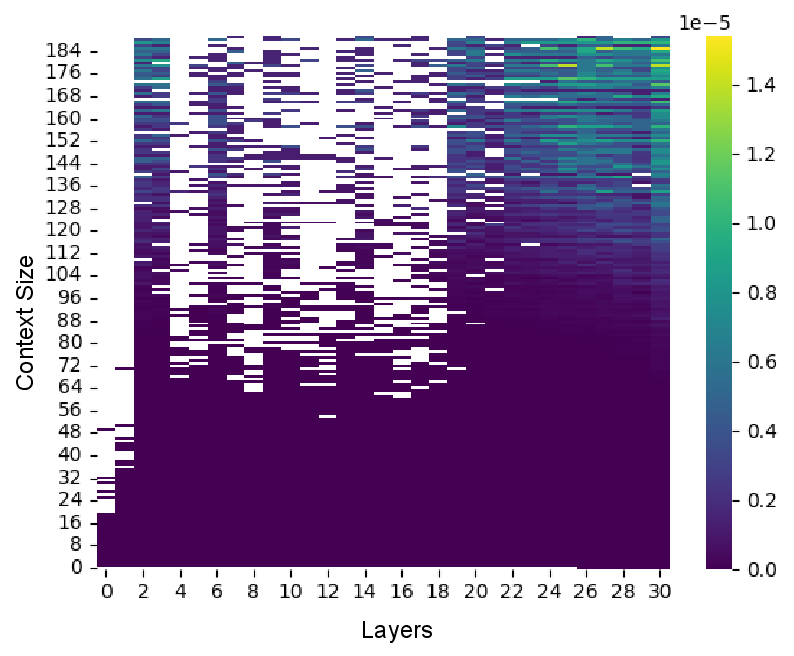}
        \caption{Go: Poly-Coder model.}
        \label{fig:sub23}
    \end{subfigure}%
    \begin{subfigure}{.45\textwidth}
        \centering
        \includegraphics[width=\linewidth]{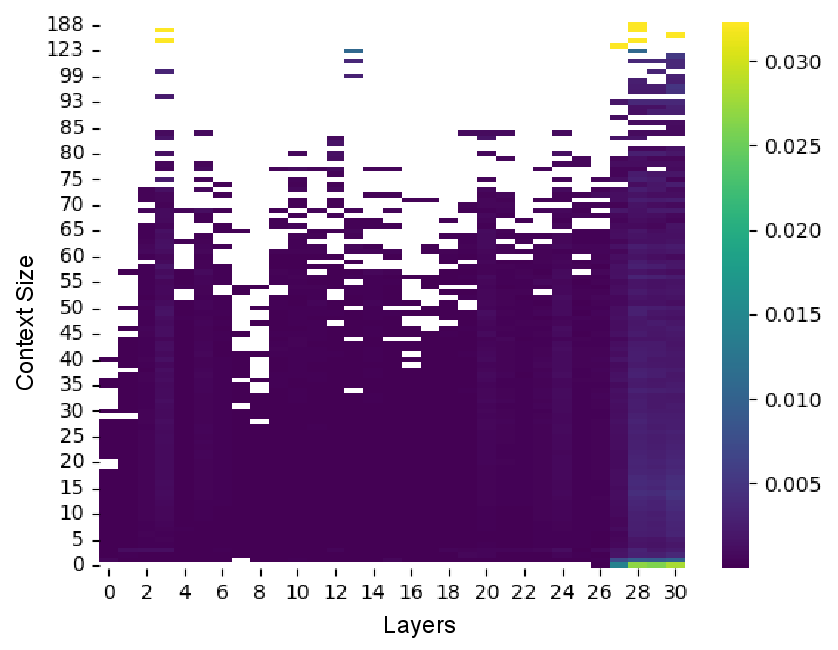}
        \caption{Java: Poly-Coder model.}
        \label{fig:sub24}
    \end{subfigure}
\caption{Layer agreements with the model's final output, as we vary the length of the context.}
\label{fig:contextsize}
\end{figure*}

To answer RQ4, how context size affects the output formulation and agreement of layers with the final output of the model, we repeat the same experiment as above with varying context sizes from 1 to 188.
We analyze the agreement among layers and token counts and present our results using 2D heatmaps in Figure~\ref{fig:contextsize}. 
We found that initial tokens are generally easier to predict, thus showing higher agreement between initial layers and the final output. This might be attributed to the model capturing more salient features in the early stages of processing.
In contrast, later tokens, which are more challenging to predict, tend to have higher agreement with the upper layers and the final output. 
This suggests that the later stages of processing, possibly involving more abstract or contextual information, play a more significant role in predicting these complex tokens.

\stitle{Python on Codegen-Mono.}
In Figure~\ref{fig:contextsize}~(a), we provide agreement results for the model Codegen-Mono on Python language of different layers with the final output of the model along with varying context sizes from 1 to 89. 
From the heatmap, we see that when the context size is small we see agreement even in the initial layers but as the context size increases only the later layers after layer 18 have agreement with the final output of the model.

\stitle{Python on Polycoder.}
Figure~\ref{fig:contextsize}~(b) presents agreement results for the model Polycoder on Python language of different layers with the final output of the model along with varying context sizes from 1 to 184. 
%
We observe similar results about the agreement, wherein as the context size increases, only the later layers exhibit agreement with the model's final output. 
However, there is a notable difference in the behavior of the agreement from Figure~\ref{fig:contextsize}~(a). 
Across all results for the Polycoder model, we observe a peak in agreement across all context sizes in the initial few layers.
After these first few layers, this behavior aligns with our observation of the agreement results for the Polycoder model (in Figure~\ref{fig:agreement1}), where the agreement increases for a few layers and then decreases after the first few layers.
We posit that this behavior stems from the inherent differences in the nature of both models. 
Nonetheless, our core assertion remains valid, as even the first layer can generate accurate predictions when the context size is relatively small.

\stitle{Go on Polycoder.}
In Figure~\ref{fig:contextsize}~(c), we provide agreement results for the model Polycoder on the Go language of different layers with the final output of the model along with varying context sizes from 1 to 184. 
We observe a similar trend to another finding in the Polycoder model, where the agreement across all context sizes initially increases around layer 5 before declining. 
However, after that, as the context size increases, only the last layers exhibit agreement with the final output of the model.

\stitle{Java on Polycoder.}
Figure~\ref{fig:contextsize}~(d) presents the agreement results of different layers within the Polycoder model, trained on the Java programming language, with the final output of the model. These agreement measurements are provided across varying context sizes, ranging from 1 to 188 tokens. 
This result is also in line with the other results of the Polycoder model model where the agreement of all context sizes increases around layer 5 and then goes down, but then as the context size increases only the last layers agree to the final output of the model.

\stitle{Findings.}
The results in Figure~\ref{fig:contextsize} help us answer RQ4: understand the behavior of the models with varying context sizes.
From these results, it is evident that the complexity of the task for the model changes with varying context sizes.
Our findings reveal that even the earlier layers, as early as the very first layer, across all four settings, can predict some tokens correctly in the smaller context size.
However, as the context size increases only the later layers can make the correct prediction except for model-dependent behavior in the results on the Polycoder model, where there was some agreement around layer 4 to layer 6 in both agreement experiments, across all settings. 
This signifies that as the context size becomes larger, the task of accurate prediction becomes difficult for the model. 
This behavior may look counter-intuitive at first because a larger context size has more information for the model to make predictions.
But a larger context also requires the model to have a higher semantic understanding of the input, which our findings from the exploration of keys suggest that only higher layers possess (refer to RQ1). 
With a smaller context size, there is a possibility that even completing n-grams and keywords could result in the correct prediction.
Our investigation into the keys has revealed that initial layers indeed demonstrate an aptitude for understanding keywords and n-grams, thereby enabling them to occasionally predict the correct output when the context size is sufficiently small.


%% file: related.tex
\section{Relater Work}
\label{sec:related}
Understanding the mechanisms behind the predictions of models is crucial for their deployment in real-world applications. Interpretability focuses on uncovering the rationale of model decisions, providing insights into model behavior, and enhancing the trustworthiness of models.
We organize related work into two categories: interpretability in machine learning and interpretability in code~LMs.

\subsection{Interpretability in Machine Learning}
The methods for achieving interpretability in machine learning models can be broadly categorized into three main types: \myNum{i}~ counterfactual interventions, \myNum{ii}~hyper-network structures, and \myNum{iii}~probing-based methods. 
The counterfactual intervention methods investigate how the changes in input features influence model outputs by modifying inputs and observing resultant output variations. 
These methods include techniques like removing or replacing input words to determine their effect on model decisions, with examples being the extraction of key sentences from labeled documents. 
The works~\cite{li2016understanding} and \cite{ribeiro2018anchors} are examples of counterfactual interventions. 
The hyper-network structure approaches involve creating a learnable mask over the neurons of a frozen pre-trained model, where an L1-norm or L2-norm is applied to the masks~\cite{haider2021comprehensive}. 
These masks serve as indicators of neuron importance in the targeted area, examples of hyper-network structure approaches are \cite{radford2019language} and \cite{lakretz2019emergence}. 
Lastly, there are probing-based methods, which involve aligning model neurons or components with specific concepts by identifying patterns of co-occurrence between neuron activations and the target concept~\cite{geva2020transformer,durrani-etal-2020-analyzing}. 
Our method of probing the model keys falls under this general category of interpretability.

Our work builds upon Geva et al.'s methodology, which pointed out that the feed-forward layers are key-value storage bases.
The domain of analysis is different, the paper of Geva et al. is in natural language processing but our work is on coding models. The definition of semantic meaning between natural language and code is very different. For example, in Geva et al.'s work examples of higher semantics are the concept of time, the concept of TV shows, etc. Whereas, in our work, we investigate how the semantics of code are formed. For example, values `0..255` being related to image (Table 4), keywords `for` and `while` being related to each other are coding concepts (Table 6). In the domain of natural language analysis, we will not find these patterns, and a genuine research question arises whether coding models contain patterns similar to natural language or a different set of coding patterns emerge. This work tries to uncover coding patterns, and their storage and retrieval mechanisms.
In the paper of Geva et al., the exploration of syntactic/semantic patterns is not targeted. That is, they randomly select keys and find trigger examples associated with those keys. Their work can not enable targeted exploration of a concept of interest, for example a specific API or method. Thus, it can not be used for eventual editing. In our work, we use regular expressions to find and analyze where information about an API of interest is stored. Due to the targeted nature of our exploration, we can leverage this information about specific concepts of interest for editing on a general API level. Moreover, we show that information is localized enough to be edited without compromising the general performance of the models.

\subsection{Interpretability in Code LMs}

Interpretability within code generation models remains a relatively under-explored area of research, and most of the research in the field focuses on the attention part of the model. Authors in \cite{mohammadkhani2023explaining} examine CodeBERT and GraphCodeBERT in the context of software engineering tasks.
By analyzing attention scores across different token types, the study reveals patterns in how these models allocate attention to various parts of the code. 
Authors in \cite{liu2024reliability} examine the effectiveness of pre-trained language models like CodeT5 and CodeGPT in generating, translating, and repairing code, They use attention interpretability specifically focusing on how these models pay attention to different parts of the code during the generation process.
Authors in\cite{paltenghi2021thinking} compare the attention mechanisms of neural models analyzing code to the attention of skilled human developers. 
It introduces a method for capturing human attention on code and compares it with the attention weights of neural models. 
To understand what coding models capture about the source code's structure and semantics, authors in \cite{wan2022they} use attention analysis along with probing on word embeddings, and syntax tree induction.
All of these works focus on analyzing attention weights and activations to understand where the model directs its attention throughout the input sequence. 
All other exploratory works in coding models focus on attention layers, which are only one-third of the model parameters. Essentially attention layers learn where the model should pay attention in the context. Previous works focus on finding out that if the model pays attention to the important coding semantic patterns in the input, it will be able to generate the correct output. These techniques fall under the umbrella of explainable AI literature where one looks to explain why the model made certain predictions. In our work, we explore the feed-forward layers, which constitute two-thirds of the model parameters, known as the databases of the model, using an interpretability method. We focus on what types of syntactic/semantic code patterns are stored in the model and identify specific neurons that store those. We then go on to show that the information stored in these layers is localized enough to be edited without a substantial impact on the model’s general coding performance. 
In summary, both types of works are complementary. Attention analysis offers valuable insights into the patterns that models prioritize and consider significant. This however does not paint the full picture of the internal workings of the model. For example, from analyzing attention, we can know which patterns are important for the model to generate correct outputs. It does not tell us how and where the information is stored, how it flows through the network to form final model predictions, which is the focus of this work. Using this knowledge, we can edit specific information from the model. In this work, we show a simple form of editing.